\documentclass[letterpaper, 10 pt, conference]{ieeeconf}  

\IEEEoverridecommandlockouts                              

\usepackage{cite}
\usepackage{amsmath,amssymb,amsfonts}
\usepackage{algorithmic}
\usepackage{subfig}
\usepackage{textcomp}
\usepackage{multirow} 

\usepackage[inkscapelatex=false]{svg}

\usepackage{graphics} 

\usepackage{epsfig} 
\usepackage{mathptmx} 
\usepackage{times} 
\usepackage{amsmath} 
\usepackage{amssymb}  
\usepackage{booktabs}
\usepackage{tabularx}

\usepackage{xcolor}
\makeatletter
\let\NAT@parse\undefined
\makeatother
\usepackage[colorlinks,citecolor=magenta,linkcolor=red,urlcolor=blue]{hyperref}

\usepackage{pifont}

\usepackage{caption}

\usepackage{colortbl}

\begin{document}

\title{\LARGE \bf
FlowPlan: Zero-Shot Task Planning with LLM Flow Engineering for Robotic Instruction Following
}

\author{Zijun Lin$^{1,2}$, Chao Tang$^{1,2,3}$, Hanjing Ye$^{1,2}$, and Hong Zhang$^{1,2}$ \emph{Life Fellow, IEEE}\\
 $^{1}$Southern University of Science and Technology, \\
 $^{2}$Shenzhen Key Laboratory of Robotics and Computer Vision, \\
 $^{3}$National University of Singapore
 \vspace{-0.6cm}
}


\twocolumn[{
\renewcommand\twocolumn[1][]{#1}
\maketitle
\begin{center}
    \captionsetup{type=figure}
    \includegraphics[width=.8\textwidth]{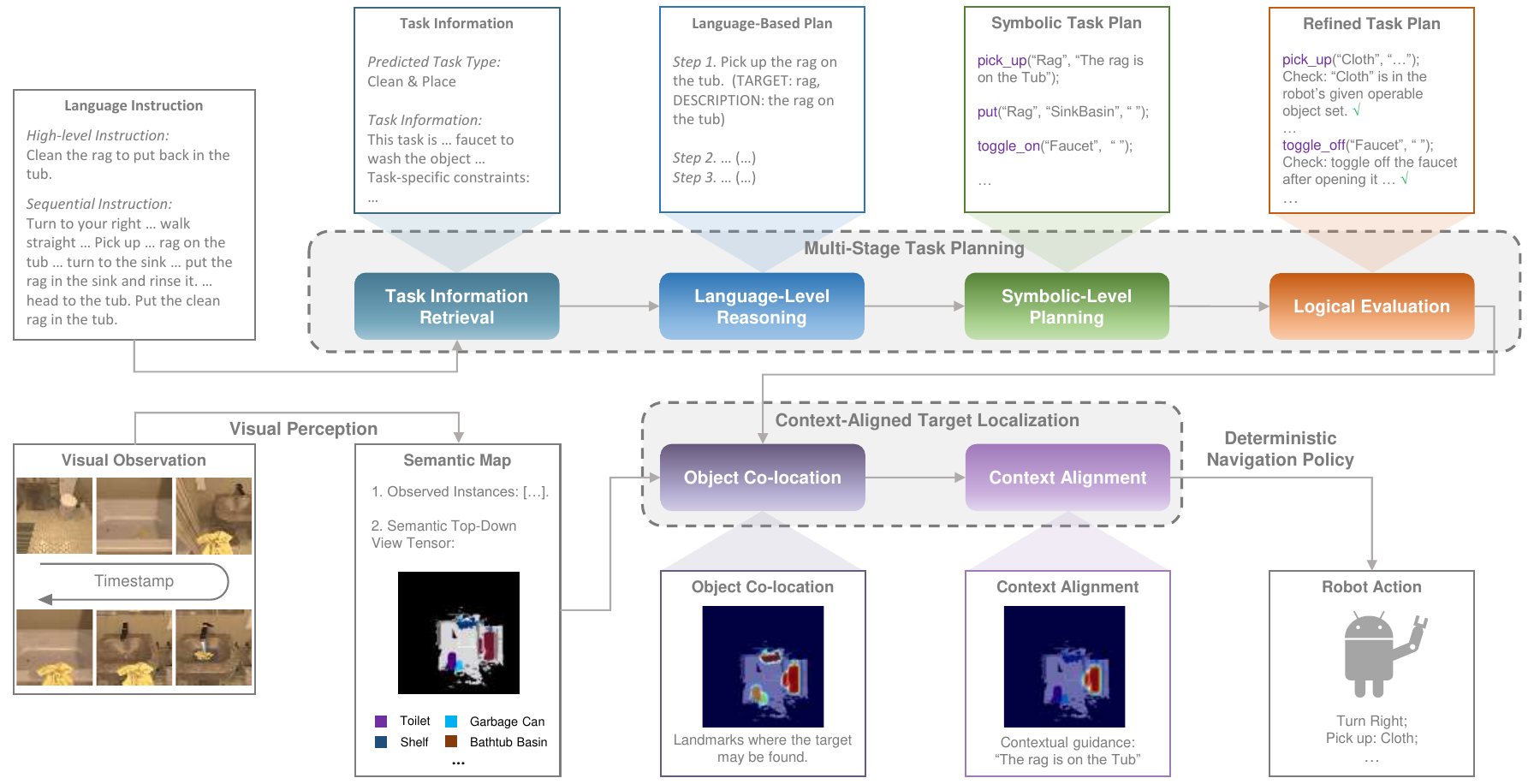}
    \captionof{figure}{ 
    \textbf{Overview}. In robotic instruction following tasks, a robot performs a series of planning steps and movements within an unfamiliar environment to achieve specific objectives. FlowPlan integrates multi-stage task planning with context-aligned target localization. The former involves task information retrieval, language-level reasoning, symbolic-level planning, and logical evaluation. The latter utilizes an online-constructed semantic map of the scene to locate targets for navigation, which predicts object co-location probabilities that are refined with contextual guidance derived from the instructions.
    }\label{fig:overview}
\end{center}
} 
]

\begin{abstract}

Robotic instruction following tasks require seamless integration of visual perception, task planning, target localization, and motion execution. However, existing task planning  methods for instruction following are either data-driven or underperform in zero-shot scenarios due to difficulties in grounding lengthy instructions into actionable plans under operational constraints. 
To address this, we propose FlowPlan, a structured multi-stage LLM workflow that elevates zero-shot pipeline and bridges the performance gap between zero-shot and data-driven in-context learning methods. By decomposing the planning process into modular stages\textemdash task information retrieval, language-level reasoning, symbolic-level planning, and logical evaluation\textemdash FlowPlan generates logically coherent action sequences while adhering to operational constraints and further extracts contextual guidance for precise instance-level target localization. Benchmarked on ALFRED and validated in real-world applications, our method achieves competitive performance relative to data-driven in-context learning methods and demonstrates adaptability across diverse environments. This work advances zero-shot task planning in robotic systems without reliance on labeled data. Project website: \href{https://instruction-following-project.github.io/}{https://instruction-following-project.github.io/}.


\end{abstract}


\section{INTRODUCTION}\label{sec: introduction}


Robotic instruction following (a.k.a. Embodied Instruction Following, EIF) tasks present significant research challenges in robotics and AI, requiring autonomous systems to perceive visual signals, interpret natural language instructions into executable task plans, locate targets for navigation, and perform physical actions. For instance, executing the command \textit{``bring me a heated apple slice"} demands sequential actions---picking up a knife, locating an apple, slicing it, placing the sliced apple in a microwave for heating, and retrieving it once the microwave has been turned off---integrated through visual perception, task planning, target localization, and physical action execution.

\begin{table*}[htbp]


  \centering

  \begin{tabularx}{\linewidth}{XX}
  \toprule


  
    \textbf{ALFRED} & \textbf{Our Application} \\

    \midrule

     (General) \textit{Single Arm Limitation: The robot cannot perform the pick\_up() action while holding an object but can still perform actions like open() and close().} & (General) \textit{Single Arm Limitation: The robot can only perform put() when holding an object.}\\

     & \\

     (General) \textit{Action Chain Integrity: Switchable objects must be turned off after use to complete the on-off action chain.}  &  (Task Specific: Heat \& Place) \textit{Cooking and Heating Restriction: The cooker automatically turns off after heating, so the robot does not need to turn it off.}  \\

    & \\

    (Task Specific: with Slicing) \textit{Slicing Restriction: The robot, limited to a single arm, must slice a target while holding the knife without picking up the target.}    &  (No Slicing is Required)  \\


    & \\

\bottomrule

\end{tabularx}
  \caption{ 
Examples of different rules and constraints in various operational contexts. These constraints contribute to the complexity of task planning, as well as pose challenges for the versatility of approaches to tackle novel scenes and tasks. The table shows some differences between the ALFRED experiments and our real-world instruction following application.}

    \label{tab:constraints}

\vspace{-0.15in}


\end{table*}

Existing approaches face significant limitations in task planning for robotic instruction following. Training-based methods~\cite{blukis2022persistent, min2021film, inoue2022prompter, kim2023context, murray2022following, bhambri2023multi, chen2023robogpt, zhao2024epo } require resource-intensive and task-specific datasets that hinder generalization. Meanwhile, data-driven in-context learning approaches~\cite{song2023llm, sarch2024helper, shi2024opex, kim2024multi} depend on manually curated labeled example sets, which still require considerable human effort in specific contexts and limit scalability. Although zero-shot solutions alleviate data dependency~\cite{shin2024socratic}, their single-stage planners struggle to (1) effectively parse lengthy instructions into logically grounded actions under operational constraints (Table~\ref{tab:constraints}), and (2) leverage contextual cues for precise target localization, resulting in only suboptimal outcomes.

When addressing various complex tasks, such as code generation or arithmetic reasoning, a deliberately designed multi-stage solution has been shown to be more effective than relying on a single prompt~\cite{ridnik2024code, aksitov2023rest, li2024survey}. For example, the authors of AlphaCodium~\cite{ridnik2024code} proposed an effective workflow that incorporates a pre-processing phase and a code iteration phase for code generation, demonstrating more promising results compared to single-stage solutions.

Inspired by multi-stage problem solving in these tasks, we propose FlowPlan: a structured LLM workflow that decomposes task planning into four interpretable stages---\textit{task information retrieval}, \textit{language-level reasoning}, \textit{symbolic-level planning}, and \textit{logical evaluation}. This modular design enables effective planning under operational constraints, as well as systematic extraction of contextual landmarks (e.g., spatial relationships) to guide target localization. As illustrated in Fig.~\ref{fig:overview}, our framework combines \textit{multi-stage task planning} with \textit{context-aligned target localization}, both implemented in a zero-shot scheme, thereby eliminating the reliance on task-specific training data.


\textbf{Contribution.} We propose FlowPlan, a multi-stage LLM workflow that elevates zero-shot pipeline and bridges the gap between zero-shot and data-driven in-context learning methods for robotic task planning. By decomposing the planning process into interpretable structured stages, our framework generates logically coherent action sequences while adhering to operational constraints and further extracts contextual guidance from instructions to enable precise target localization. Experiments on the ALFRED benchmark demonstrate significant performance improvements in zero-shot scenarios, with real-world deployments further validating its versatility across diverse environments and tasks.

    


\section{Related Work}

\subsection{Task Planning with LLMs}

Recent advances in Large Language Models (LLMs) have enabled robots to interpret natural language instructions and generate action plans through LLM-based reasoning~\cite{chen2023llmstate,rana2023sayplan, ahn2022can,chen2023autotamp,brohan2022rt,brohan2023rt,jiang2022vima, singh2023progprompt}. While they are promising for simplified tasks characterized by short and straightforward instructions or under ideal perception and target localization conditions, complex EIF tasks~\cite{shridhar2020alfred}, which require robots to translate lengthy instructions and achieve objectives within intricate interactive scenarios, still remain a significant challenge. Most existing EIF solutions are data-driven and can be categorized into two main types: (1) Training-based approaches~\cite{blukis2022persistent, min2021film, inoue2022prompter, kim2023context, murray2022following, bhambri2023multi, chen2023robogpt, zhao2024epo} that require extensive human-annotated datasets, thereby limiting their generalization capabilities, and (2) Data-driven in-context learning methods~\cite{song2023llm,shi2024opex,sarch2024helper, kim2024multi} which reduce data dependency but demand laborious manual curation of example sets for similarity-based retrieval.
For instance, in~\cite{song2023llm,shi2024opex}, the authors manually create an example set and develop algorithms to extract the top-k examples based on linguistic similarity for the purpose of in-context learning.

To alleviate this burden, Shin et al.~\cite{shin2024socratic} endeavored to address the task through a zero-shot approach. However, their single-stage planners struggle to effectively parse lengthy instructions into accurate task plans due to the ambiguous expression of the instructions and the underlying operational constraints, as discussed in Section~\ref{sec: introduction}. This limitation results in suboptimal performance and highlights a significant gap in the development of data-efficient frameworks that are capable of effectively addressing the instruction following problem.


\begin{figure*}[htbp]
    \centering
    \includegraphics[width=1.0\linewidth]{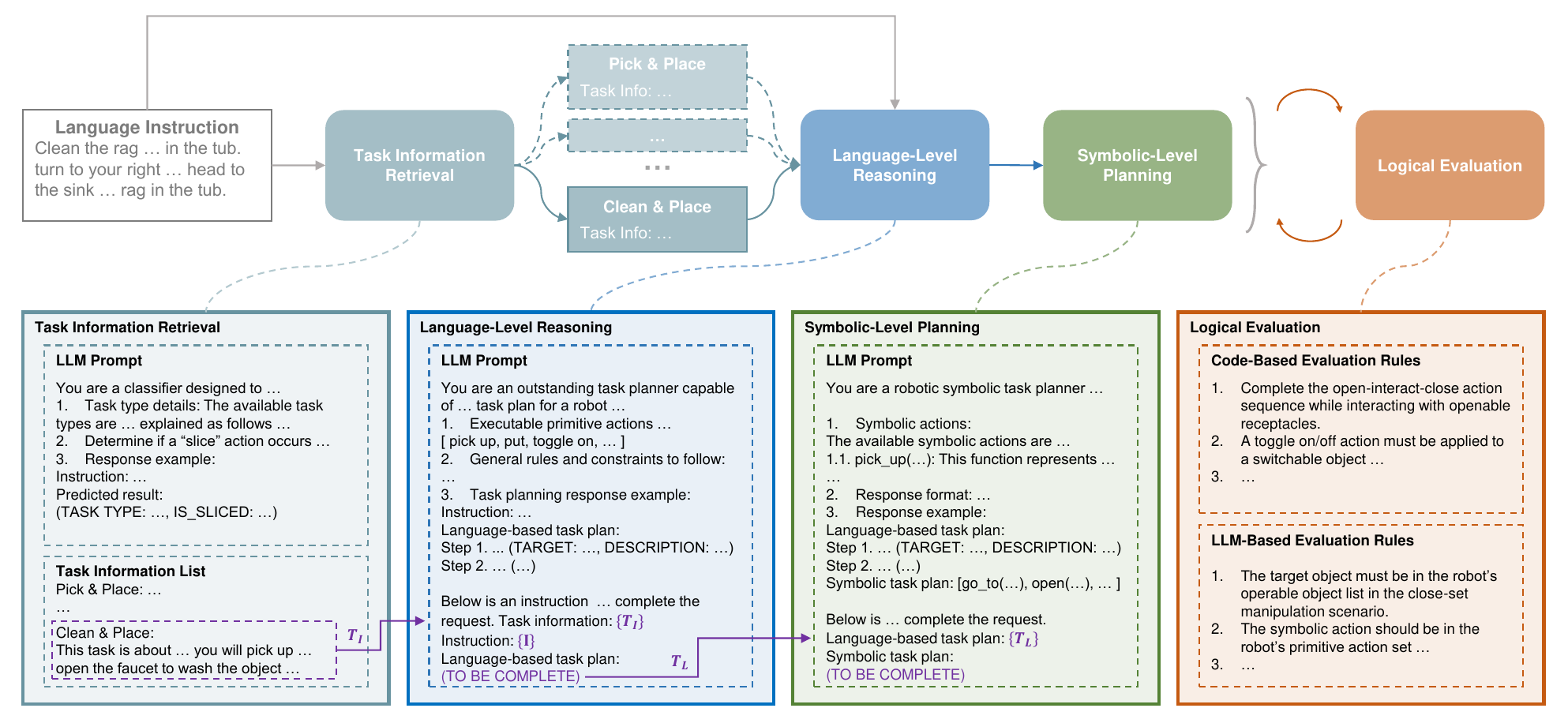}
    \caption{\label{fig: CTP module.} 
    \textbf{Multi-Stage Task Planning.} The multi-stage task planning process is comprised of four interpretable stages\textemdash task information retrieval, language-level reasoning, symbolic planning, and logical evaluation\textemdash to generate logically coherent task plans under operational constraints. All stages are managed by LLM-driven components and do not require labeled data or example sets.
    } 
    \vspace{-0.15in}
\end{figure*}

\subsection{Prompt Chaining and Multi-Stage LLM Workflows}

Prompt chaining and multi-stage LLM workflows have been extensively studied and demonstrated to be effective in addressing various complex reasoning and generation challenges~\cite{aksitov2023rest, ridnik2024code, yin2024novel, yao2024hdflow, han2025multi, trirat2024automl, grunde2023designing, xiao2024flowbench, fan2024workflowllm, li2024autodcworkflow}. For instance, Ridnik et al.~\cite{ridnik2024code} introduced an effective workflow for code generation that comprises a pre-processing phase and a code iteration phase. The results indicate that this approach outperforms a well-structured single prompt. 
Li et al.~\cite{li2024autodcworkflow} implemented a pipeline for an LLM-based automated data cleaning workflow that contains three main LLM-driven components. Similarly, Yao et al.~\cite{yao2024hdflow} proposed a dynamic workflow that automatically decomposes complex problems into more manageable subtasks, significantly outperforming Chain-of-Thought (CoT) reasoning. Inspired by these works, we aim to explore how a deliberately designed multi-stage LLM workflow can enhance task planning and target localization.


\section{Method}


Our framework enhances zero-shot task planning for instruction following through a structured pipeline comprising two components: (1) \textit{multi-stage task planning} that translates instructions into action-target pairs under operational constraints while extracting contextual guidance from the instructions, and (2) \textit{context-aligned target localization} that locates targets in the scene with grounded contextual guidance to facilitate subsequent navigation. The architecture is detailed in Fig.~\ref{fig:overview}.

\subsection{Multi-Stage Task Planning}


Given a language instruction \( I \), the task planner \( f_{TP}: I \rightarrow \{T_n\}_{n=1}^N \) generates executable action sequences, where each step \( T_n = (A_n, O_n, C_n) \) contains:

\begin{itemize}
\item \( A_n \): Primitive action (e.g., \textit{open}),
\item \( O_n \): Target object label (e.g., \textit{cabinet}),
\item \( C_n \): Contextual guidance for instance grounding (e.g., \textit{the cabinet beneath the coffee machine}).
\end{itemize}

As shown in Fig.~\ref{fig: CTP module.}, the planning pipeline implements four cascaded stages:

\textbf{Task Information Retrieval.} As illustrated in~\cite{shridhar2020alfred}, language instructions can be classified into various task categories, such as simple pick-and-place operations and heating an object in a microwave. Understanding the type of task is beneficial for LLMs as it enhances their comprehension of the instructions. In this study, we identify seven distinct task types: \textit{\{Pick \& Place, Stack \& Place, Clean \& Place, Cool \& Place, Heat \& Place, Pick Two \& Place, Examine in Light\}}. We also classify tasks that involve slicing objects independently. Each type of task is associated with task information \( T_I \), which includes a concise description that captures the essence of the task, along with pertinent operational rules and constraints (as described in Table~\ref{tab:constraints}), all of which can be easily articulated in natural language without much effort.

Upon receiving an instruction, an LLM is initially tasked with categorizing the instruction, which predicts the fundamental task type. The accuracy of this task type prediction is further validated through a voting mechanism among the top-k responses generated by the LLM or by employing evaluations from multiple LLM calls. Following this categorization, the relevant task information \( T_I \) is retrieved and incorporated into subsequent stages via templated prompts.

\textbf{Language-Level Reasoning.} 
A reasoning step is then conducted to transform instruction \(I\) into a language-level task plan \(T_L\) with the awareness of task information \(T_I\).
In previous work (e.g.,~\cite{shin2024socratic, song2023llm}), the authors directly translated instructions into symbolic-level plans without explicit language-level reasoning. This approach has been shown to negatively impact general LLMs due to logical and grammatical ambiguities (e.g., implicit state transitions) and the underlying operational constraints (Table~\ref{tab:constraints}) in the instructions.
Although data-driven in-context learning methods (e.g.,~\cite{min2021film,inoue2022prompter,kim2023context,shi2024opex}) can alleviate these issues to some extent by learning from examples of similar tasks, they remain data-dependent and may still experience confusion at times.
Therefore, in this study, we propose to first conduct a preliminary stage that preprocesses the instructions and reasoning at the natural language level prior to symbolic planning, which is advantageous in zero-shot scenarios.
To illustrate this with an example, for the instruction \textit{``pick up the apple in the drawer and ..."}, an explicit reasoning step at the language level could complete the logically coherent sequence: \textit{\{``1. Open the drawer $\rightarrow$ 2. Pick up the apple $\rightarrow$ 3. Close the drawer"\}}.

To better achieve the aforementioned language-level reasoning under constraints, this component is designed as an independent LLM-driven component tasked with enhancing clarity, facilitating reasoning, and translating instructions into primitive action steps at the language level. This component operates under the direction of a prompt that outlines its role, specifies executable primitive actions (including \textit{``pick up", ``put", ``toggle on", ``toggle off", ``open", ``close", ``slice", ``go to a landmark"}), and defines general operational rules and constraints. Additionally, it provides a response example to assist in the generation process of the LLM.

The task information retrieved in the previous step will be synthesized to formulate a task-aware prompt. The operational rules and constraints specific to the tasks can be utilized to mitigate ambiguity in the instructions and to generate logically coherent sequences of actions. For instance, if the task pertains to actions involving the slicing of an object, a specific slicing restriction, as demonstrated in Table~\ref{tab:constraints}, will be provided to the robot. Without an understanding of these operational constraints, LLMs may reasonably deduce that the slicing action requires \textit{``picking up the object and slicing it"}, which might be plausible for a human in certain situations but is inconsistent with the robot's operational constraints.

\textbf{Symbolic-Level Planning.} Following the language-level reasoning, a subsequent symbolic-level planning phase is conducted to generate an executable symbolic action sequence \( \{ T_{n} \}_{n=1}^{N} \) based on the language-based task plan \( T_L\). In addition to the target object label \(O_{n}\) and action \( A_{n} \), as noted in~\cite{min2021film,inoue2022prompter,kim2023context}, the substep of our approach also incorporates relevant contextual guidance information \( C_{n} \) (e.g., \textit{``the cabinet beneath the coffee machine"}) extracted from the earlier stage to facilitate the subsequent target localization process.

The implementation of the LLM prompt consists of several components: a role description that outlines its responsibilities and objectives; symbolic actions that represent high-level actions that can be directly executed by the robot; a description of the response format; and an example response to guide the generation of the symbolic task plan.


\textbf{Logical Evaluation.}
Upon generating the task plan, an evaluation stage is implemented, which includes both code-based and LLM-based evaluations. The code-based evaluation is responsible for verifying logical coherence and operational constraints. The rules and operational constraints, as described in Table~\ref{tab:constraints}, will be provided as a straightforward configuration that describes valid, invalid, and fixed action sequences. This configuration is designed to be easily adaptable across various operational contexts with minimal effort. For instance, the first constraint for the Franka robot in our real-world application, as shown in Table~\ref{tab:constraints}, can be succinctly described as a fixed action sequence \{\textit{pick\_up() $\rightarrow$ put()}\}, where any other actions with an object in hand \{\textit{pick\_up() $\rightarrow$ action()}\} will be considered invalid. The LLM-based evaluation, on the other hand, primarily addresses the correction of invalid items in the task plan. In our close-set manipulation setting, when a singular action or object label term appears in the task plan, the LLM will first attempt to replace it with the most similar items from the valid operational object and action list, which resembles lexical matching.

If the evaluation result shows that the task plan does not conform to the logical constraints or still contains invalid items after several correction attempts, it will initiate a re-planning phase, which involves re-conducting the task planning process. Unlike scene-aware approaches~\cite{kim2023context,song2023llm,shin2024socratic} that re-plan after failures and leverage previous scene observations, our study focuses on task planning, excluding scene-aware re-planning.

\subsection{Context-Aligned Target Localization}



The target localization process is tasked with locating targets in the scene for subsequent navigation actions. The process \( f_{GP}: (T_n, S) \rightarrow G \) maps the current subtask \( T_n \) and an online-constructed semantic map \( S \) to a navigation goal \( G \). Here, \( S = \{ M, L \} \) comprises:  
\begin{itemize}
    \item \( M \): \( W \times W \times C \) semantic top-down view tensor constructed from depth prediction and semantic segmentation~\cite{min2021film}, and
    \item \( L \): Observed instance coordinates with categorical labels.
\end{itemize}

\begin{figure}[t]
    \centering
    \includegraphics[width=1.0\linewidth]{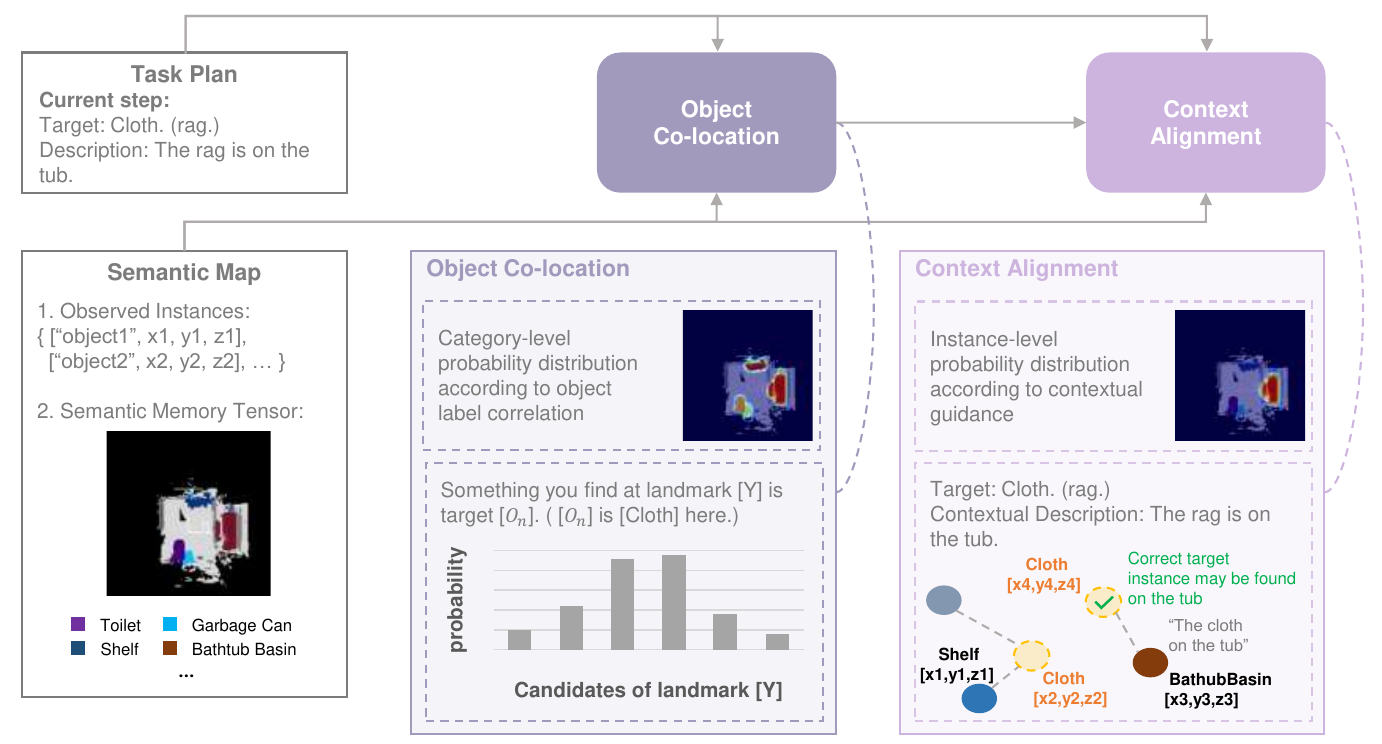}
    \caption{\label{fig: CSP module}\textbf{ 
    Context-Aligned Target Localization.} The target localization process consists of two key components: object co-location and context alignment. The former produces a probability distribution at the category level, while the latter utilizes guidance from instructions to align with a specific target instance.} 
    \vspace{-0.1in}
\end{figure}

As shown in Fig.~\ref{fig: CSP module}, our zero-shot target localization pipeline combines two complementary stages:


\textbf{Object Co-location.} 
Unobserved targets are localized via categorical relationships (e.g., \textit{knife} → \textit{kitchen counter}). We first compute a category-level probability map \( D_o \) through linguistic correlation analysis, following a similar process described in~\cite{inoue2022prompter}. Given the target object \( O_n \), the LLM predicts object co-location probabilities between \( O_n \) and observed landmarks \( Y \in L \). The predicted grid values are aggregated along the category channels to produce a \( W \times W \) probability map for navigation.

\begin{table*}[htbp]
  \centering

  \begin{tabularx}{\textwidth}{lccXXXXXXXX}
  \toprule
     & &  & \multicolumn{4}{c}{Test Seen} & \multicolumn{4}{c}{Test Unseen} \\ 
    
         \cmidrule(lr){4-7} \cmidrule(r){8-11}
    
        Method & Labeled Data Requirement & Data-Free & SR & GC & PLWSR & PLWGC  & SR & GC & PLWSR & PLWGC  \\ 
    
    \midrule
    
    \multicolumn{6}{l}{Training-based (using thousands of instrution--ground truth data pairs for training)} & & & & &  \\

    \midrule
    
    \textcolor{gray}{FILM~\cite{min2021film}} & \textcolor{gray}{training}  &  \color{red}\ding{56}   & \textcolor{gray}{27.67} & \textcolor{gray}{38.51} & \textcolor{gray}{11.23} & \textcolor{gray}{15.06} & \textcolor{gray}{26.49} & \textcolor{gray}{36.37} & \textcolor{gray}{10.55} & \textcolor{gray}{14.30} \\
    
    \textcolor{gray}{LGS - RPA~\cite{murray2022following}}   & \textcolor{gray}{training}  & \color{red}\ding{56}  & \textcolor{gray}{40.05} & \textcolor{gray}{48.66} & \textcolor{gray}{21.28} & \textcolor{gray}{28.97} & \textcolor{gray}{35.41} & \textcolor{gray}{45.24} & \textcolor{gray}{15.68} & \textcolor{gray}{22.76} \\
    
    \textcolor{gray}{Prompter~\cite{inoue2022prompter}} & \textcolor{gray}{training}  & \color{red}\ding{56}  & \textcolor{gray}{51.17} & \textcolor{gray}{60.22} & \textcolor{gray}{25.12} & \textcolor{gray}{30.21} & \textcolor{gray}{45.32} & \textcolor{gray}{56.57} & \textcolor{gray}{20.79} & \textcolor{gray}{25.8} \\
    
    \textcolor{gray}{CAPEAM~\cite{kim2023context}}  & \textcolor{gray}{training}  & \color{red}\ding{56}  & \textcolor{gray}{51.79} & \textcolor{gray}{60.50} & \textcolor{gray}{21.60} & \textcolor{gray}{25.88} & \textcolor{gray}{46.11} & \textcolor{gray}{57.33} & \textcolor{gray}{19.45} & \textcolor{gray}{24.06} \\
    
    \textcolor{gray}{MCR - Agent~\cite{bhambri2023multi}}  & \textcolor{gray}{training}  & \color{red}\ding{56}  & \textcolor{gray}{30.13} & \textcolor{gray}{$N/A$} & \textcolor{gray}{21.19} & \textcolor{gray}{$N/A$} & \textcolor{gray}{17.04} & \textcolor{gray}{$N/A$} & \textcolor{gray}{9.69} & \textcolor{gray}{$N/A$} \\
    
    \textcolor{gray}{EPO~\cite{zhao2024epo}} & \textcolor{gray}{training}  & \color{red}\ding{56}  & \textcolor{gray}{64.79} & \textcolor{gray}{72.30} & \textcolor{gray}{56.92} & \textcolor{gray}{66.20} & \textcolor{gray}{62.35} & \textcolor{gray}{67.52} & \textcolor{gray}{51.99} & \textcolor{gray}{64.15} \\

    \midrule
    \multicolumn{8}{l}{Data-driven in-context learning (using dozens of instruction--ground truth data pairs for retrieval)} & & &   \\
    \midrule

    LLM-Planner~\cite{song2023llm} & example set &  \color{red}\ding{56} & 18.20 & 26.77 & $N/A$ & $N/A$ & 16.42 & 23.37 & $N/A$ & $N/A$ \\  

    
    HELPER-$X_P$~\cite{sarch2024helper}   & example set  &  \color{red}\ding{56}  & 28.2 & 42.5 & 7.5 & 12.9 & 35.4 & 52.9 & 7.9 & 12.3 \\ 

    OPEx~\cite{shi2024opex}  & example set &  \color{red}\ding{56} & \textbf{44.03} & \textbf{54.81} & 14.52 & \textbf{22.08} & \textbf{41.85} & \textbf{54.18} & 13.48 & 15.27 \\

    FLARE~\cite{kim2024multi} & example set &  \color{red}\ding{56} & 40.05 & 48.84 & \textbf{16.68} & 21.31 & 40.88 & 51.72 & \textbf{18.14} & \textbf{22.78} \\

    \midrule
    \multicolumn{6}{l}{Zero-shot setting (no labeled data requirement)} & & & & &  \\

    \midrule

    LLM-Planner*~\cite{song2023llm} & no requirement &  \color{green}\ding{52}  & 5.71 & 10.42 & $N/A$ & $N/A$ & 6.54 & 12.12 & $N/A$ & $N/A$ \\  

    Socratic-Planner~\cite{shin2024socratic}  & no requirement  &  \color{green}\ding{52}  & 12.26 & 20.86 & $N/A$ & $N/A$ & 9.94 & 18.88 & $N/A$ & $N/A$ \\ 

    \rowcolor{gray!30}
    \textbf{FlowPlan (Ours)} & no requirement &  \color{green}\ding{52}  & \textbf{40.31} & \textbf{49.35} & \textbf{16.31} & \textbf{19.71} & \textbf{35.64} & \textbf{46.93} & \textbf{14.06} & \textbf{17.89} \\

  \bottomrule
  
  \end{tabularx}

    \caption{
Quantitative comparison on ALFRED\cite{shridhar2020alfred}. Experiments indicate that our method outperforms previous zero-shot approaches and demonstrates competitive performance when compared to data-driven in-context learning approaches. Additionally, we include results reported by trained methods (indicated in \textcolor{gray}{gray text}) for reference. For models that did not report goal-conditioned or path length weighted metrics, we noted ``$N/A$" in the comparison. The results marked with an asterisk (*) represent outcomes produced by~\cite{shin2024socratic}.}

    \label{tab:comp-quant-result}

  \vspace{-0.15in}

\end{table*}

\textbf{Context Alignment.}
When contextual guidance \( C_n \) exists (e.g., \textit{``the cabinet beneath the coffee machine"}), an instance-level distribution \( D_c \) is generated. This process is similar to previous research~\cite{dorbala2023can} that enables navigation to uniquely described target objects in previously unseen environments with LLMs and VLMs (vision-language models) in the context of object navigation~\cite{chaplot2020object, ramakrishnan2022poni, chang2023goat, luo2022stubborn}. We implement a relatively simpler module that incorporates a zero-shot LLM-driven component to demonstrate the application of the extracted contextual guidance in locating target objects. The LLM operates with an awareness of the current subtask \( T_n \) and the observed instance list \( L \). It identifies the optimal instance from \( L \) by aligning the spatial relationships of \( C_n \) and subsequently outputs the target coordinates \( (x,y) \). These coordinates are then transformed into a Gaussian distribution \( D_c \) centered at \( (x,y) \).

Subsequently, the two distributions \(D_o\) and \(D_c\) are integrated through multiplication and normalization to form a unified distribution, from which a final two-dimensional navigation goal is sampled. A deterministic navigation approach, as illustrated in~\cite{inoue2022prompter, murray2022following}, is then employed to facilitate motion planning and achieve the objective, followed by the execution of the target actions associated with the substeps.

\section{Experiments}

\subsection{Experiments on the ALFRED Benchmark}

\textbf{Benchmark.} To demonstrate the effectiveness of our method, we employ the ALFRED benchmark~\cite{shridhar2020alfred}, which presents a challenging environment for robotic instruction following tasks. The ALFRED test set is categorized into two segments: ``Test Seen" (1,533 episodes) and ``Test Unseen" (1,529 episodes). The terms ``Seen" and ``Unseen" indicate whether the corresponding environments were included in the training dataset. It is noteworthy that our zero-shot pipeline functions without prior training on the training set. Furthermore, the validation set is segmented into ``Valid Seen" (820 episodes) and ``Valid Unseen" (821 episodes), which we employed for conducting ablation studies and qualitative analysis.

\textbf{Metrics and Evaluation.} The metrics include Task Success Rate (SR), Goal-Conditioned Success (GC), and their path length weighted versions, PLWSR and PLWGC, as described in~\cite{shridhar2020alfred}. The SR measures the percentage of tasks completed by the robot, while GC quantifies the ratio of achieved goal conditions. PLWSR and PLWGC are calculated by weighting SR and GC according to the ratio of the expert trajectory length to the robot's path length.

\textbf{Base Model and Prompting.} Our pipeline is capable of incorporating any LLM as the base model. For a fair comparison, we utilize the public GPT-4~\cite{achiam2023gpt} (\texttt{GPT-4-0125-preview}) as the base model, with a maximum token limit of $4,096$ and a temperature setting of $1$ to ensure appropriate re-generation in the logical evaluation-correction loop. Although a single example is provided in the prompt to guide the LLM's response in the correct format, our method does not require labeled data, thereby maintaining a zero-shot pipeline.


\begin{figure}[t]
    \centering
    \includegraphics[width=1.0\linewidth]{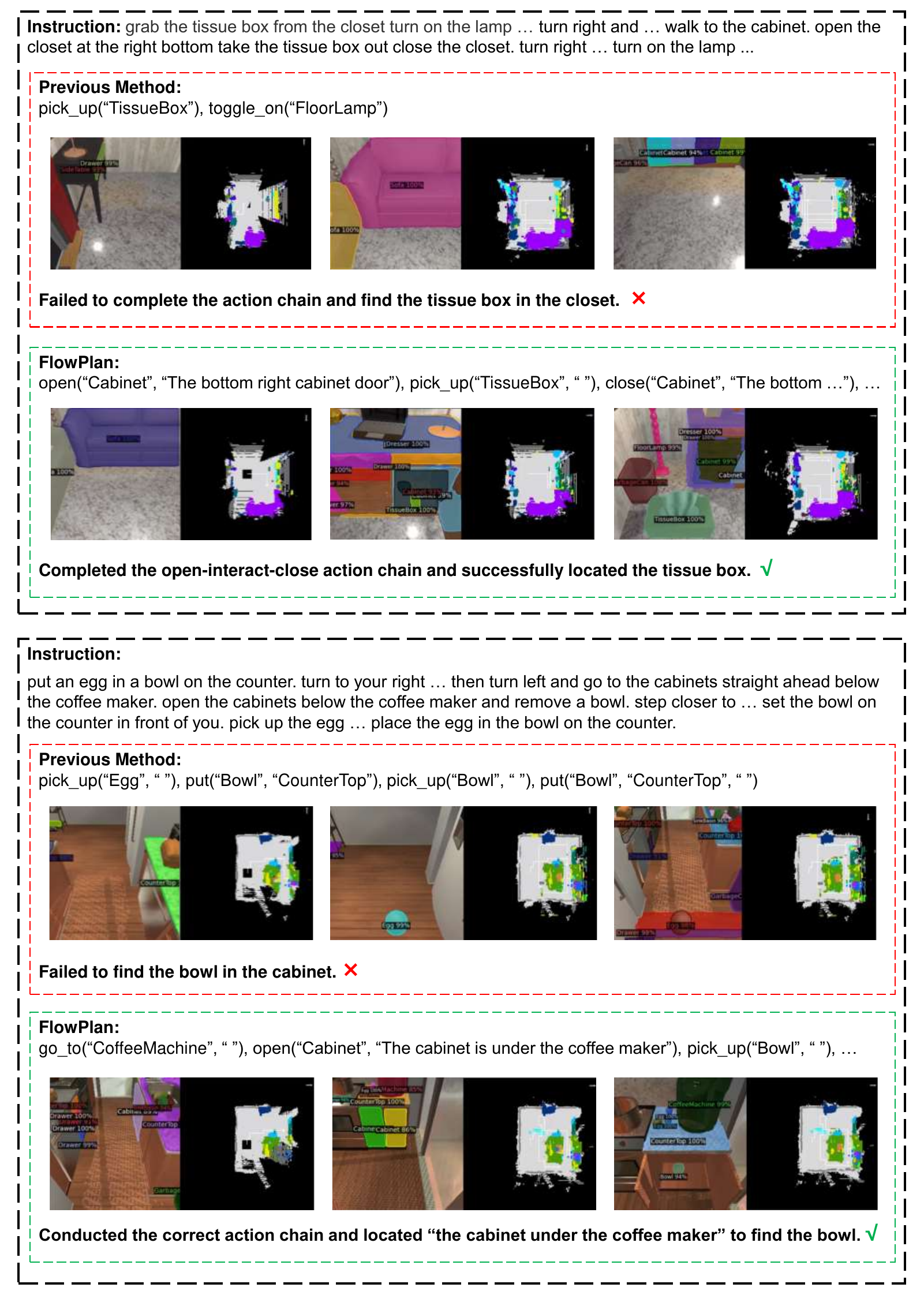}
    \caption{\label{fig: visualization} 
    \textbf{Visualization of Action Sequences.} Our method generates logically coherent task plans and executes them effectively, whereas Inoue's method~\cite{inoue2022prompter} fails.}
    \vspace{-0.15in}
\end{figure}

\subsection{Baselines and Quantitative Comparison}

Our experiments demonstrate significant advancements in the zero-shot scenario, as illustrated in Table~\ref{tab:comp-quant-result}. 
Among the methods utilizing a zero-shot pipeline, FlowPlan exhibits remarkable superiority, achieving an SR of $40.31\%$ and a GC of $49.35\%$ on the ``Test Seen" set, as well as an SR of $35.64\%$ and a GC of $46.93\%$ on the ``Test Unseen" set, with the latter being slightly lower due to lower visual perception quality. This performance is twice that of the best zero-shot competitor~\cite{shin2024socratic}, which recorded $12.26\%$ SR and $20.86\%$ GC on ``Test Seen", as well as $9.94\%$ SR and $18.88\%$ GC on ``Test Unseen", respectively. This substantial improvement underscores the efficacy of our multi-stage solution in translating complex instructions into grounded actions. 

Notably, even in the absence of labeled data, FlowPlan remains competitive with data-driven in-context learning methods that require data retrieval from labeled example sets~\cite{song2023llm, sarch2024helper, shi2024opex, kim2024multi}. It achieves an SR that surpasses some of these approaches and is only slightly lower than the highest score in the split, with a margin of just $3.72\%$ SR on the ``Test Seen" set and a slightly larger margin of $6.21\%$ SR on the ``Test Unseen" set. Although the approaches that necessitate re-training models~\cite{min2021film, murray2022following, inoue2022prompter, kim2023context, bhambri2023multi, zhao2024epo} are not the primary focus of our comparison, we have included their results for reference in Table~\ref{tab:comp-quant-result}.

\subsection{Ablation Study}

To gain a deeper understanding of the significance of each component, we conduct ablation studies. With symbolic-level planning and object co-location serving as fundamental components, the ablation is across several dimensions: the base method, the method excluding task-aware information, language-level reasoning, logical evaluation, and contextual guidance. Table~\ref{tab:ablation-result} presents the ablation results.

The performance experiences a significant decline (over \( 50 \%\) of the precision drop in SR) in both the seen and unseen sets when the logical evaluation stage is omitted. This underscores the importance of refining the task plan with mandatory constraints, valid actions, and operable object sets. The length and inherent ambiguity of the instructions often lead to flawed task plans, making the evaluation and correction step crucial for enhancing performance. The impact of reasoning at the language level is also substantial, with approximately a \(30 \%\) decrease in SR when conducting symbolic planning directly. Generating a symbolic action sequence in such a complex task scenario clearly increases the burden on a single LLM call, resulting in less accurate outcomes. We then experimented by removing the task information. The results indicate that knowledge of task-specific information, along with relevant rules and constraints, is beneficial for LLMs in generating a more effective task plan. The impact of aligning contextual guidance is relatively incremental, with only a small portion of the instructions in ALFRED~\cite{shridhar2020alfred} providing meaningful spatial relationship guidance that can be utilized to improve target localization.

\subsection{Qualitative Analysis and Discussion}

To achieve a more comprehensive understanding of the robot's behavior, we conduct a qualitative analysis. Fig.~\ref{fig: visualization} illustrates the comparison between our approach and the previous method~\cite{inoue2022prompter}. While the robot in~\cite{inoue2022prompter} encountered failures in generating the correct action chain, our robot successfully adheres to the instructions and operational constraints to complete the accurate action chains. 

\begin{table}[t]

  \centering

  \begin{tabularx}{\columnwidth}{lXXXX}
  \toprule

     & \multicolumn{2}{c}{Valid Seen} & \multicolumn{2}{c}{Valid Unseen} \\

    \cmidrule{2-3} \cmidrule{4-5}
  
    Method &  SR & GC & SR & GC  \\

    \midrule
    
    w/o Task Information   & 39.15  & 50.12  & 37.03 &  48.11 \\
    
    w/o Language-Level Reasoning   & 30.73 & 39.88 & 26.31 & 36.05 \\

    w/o Logical Evaluation    &  19.63 & 28.78 & 20.10 & 28.56 \\

    w/o Contextual Guidance    & 43.05 & 54.33 & 41.66 & 52.38   \\ 

    \midrule

    \textbf{Base Method}  & \textbf{44.15} & \textbf{57.07}  & \textbf{42.87} & \textbf{54.52}  \\

  \bottomrule
  \end{tabularx}
    \caption{Ablation study on the validation set.}
      \label{tab:ablation-result}
  
\end{table}

\begin{table}[tbp]

  \centering

  \begin{tabularx}{\columnwidth}{lX}
  \toprule

    \multicolumn{2}{c}{Close-Set Manipulation Task}  \\ 

    \midrule
    Objects & \textit{\{cooker, drawer, countertop, pan, bowl, lemon, orange, paper box, strawberry, brown cube, tissue pack\} } \\

     & \\

    Primitive Actions & \textit{ \{pick\_up(), put(), open(), close(), toggle\_on()\} } \\

     & \\

    Task Categories & \textit{ \{Pick \& Place, Heat \& Place, Stack \& Place\} }\\

    \bottomrule
\end{tabularx}

  \caption{Real-world instruction following task settings.}
  \label{tab:real-world}

  \vspace{-0.15in}

\end{table}

We also examine instances in which the robot fails to complete tasks and identify potential underlying causes. Perception failures, as discussed in previous studies~\cite{min2021film, inoue2022prompter, kim2023context}, remain a significant factor contributing to task failure. This issue could potentially be mitigated through the implementation of more effective semantic segmentation and depth prediction models, as well as a larger field of view in real applications; however, these are not the primary focus of our research.

The LLM-driven task planner exhibits occasional failures when confronted with ambiguous expressions in instructions in the ALFRED benchmark~\cite{shridhar2020alfred}. For instance, the phrase \textit{``the green cup"} may be expressed as \textit{``the green bin"} within the instructions, leading the LLM task planner to erroneously categorize it as a \textit{``trash bin"}. Addressing this issue necessitates the integration of enhanced visual reasoning capabilities into the pipeline. Such cases contribute to a proportion of failures.

The context alignment module may occasionally experience failures attributed to inaccurate translation of instructions or deficiencies in instance-level alignment. Nevertheless, such failures do not adversely affect overall performance, as the suboptimal predictions\textemdash despite their limitations in accurately reasoning through instance-level information\textemdash remain within an acceptable distribution range corresponding to the relevant object category.

\subsection{Real-World Robotic Instruction Following Experiments}

\begin{figure}[t]
    \centering
    \includegraphics[width=1.0\linewidth]{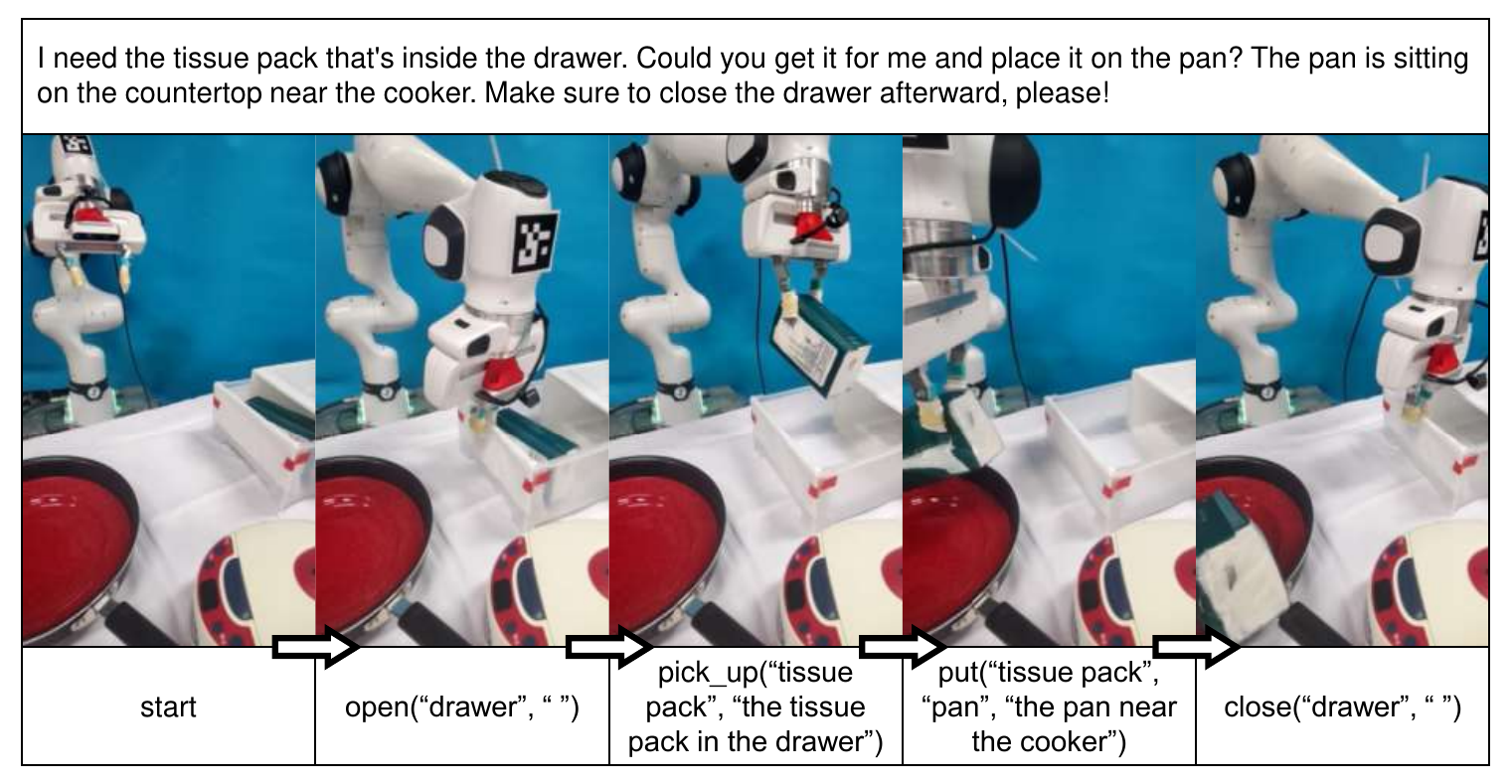}
    \caption{\label{fig: real-world}Real-world instruction following experiments.} 
\end{figure}

\begin{table}[t]

  \centering
  \begin{tabularx}{0.75\columnwidth}{XX}
  \toprule
    Method & Task Planning Precision \\

    \midrule

    Vanilla LLM & 0.76 (38/50) \\
    FlowPlan & 1.00 (50/50) \\

\bottomrule

\end{tabularx}

  \caption{Task planning results of the real-world instruction following experiments.}
  \label{tab:real-world-result}
  \vspace{-0.15in}
\end{table}

Our pipeline can be readily applied to novel scenes and tasks. In contrast to previous data-driven approaches~\cite{min2021film,inoue2022prompter,kim2023context,murray2022following, chen2023robogpt,zhao2024epo,song2023llm,sarch2024helper,shi2024opex } that necessitate the re-training of models or the reconstruction of labeled example sets for new scenarios, our pipeline requires only a few sentences of natural language descriptions regarding fundamental operational constraints. This significantly reduces the effort required from human operators.
We applied our pipeline in a real-world instruction following application utilizing a Franka Emika FR3 platform, as shown in Fig.~\ref{fig: real-world}. The vision sensor affixed to the robotic arm is an Intel RealSense D435 camera, which has undergone eye-in-hand calibration.
Close-set manipulation is conducted utilizing valid operable objects, primitive actions, and task categories as outlined in Table~\ref{tab:real-world}. We use LLM to generate a comprehensive set of instructions comprising 50 natural, conversational, and human-like casual directives tailored to the specified settings. Additionally, synonyms for object terms are incorporated into a portion of the instruction set.

Due to the diverse and intricate nature of manipulation actions, the RGB observation of the scene is initially segmented using the Segment Anything Model (SAM)~\cite{kirillov2023segment}, and the target interaction masks for the actions are provided to the robot. During the execution of each step in a task plan, the target point clouds are derived from RGB-D sensor data and segmentation results, which are subsequently utilized to generate end effector poses. For the grasping action, we employ Contact-GraspNet~\cite{sundermeyer2021contact} to generate 50 candidate grasp poses. These candidates are then filtered and ranked based on the angles between the poses and the horizontal plane, as well as their distances to the target center.

The results of task planning are presented in Table~\ref{tab:real-world-result}. For comparative purposes, we employ the Vanilla LLM, which is structured as a single, well-defined prompt. In this relatively uncomplicated task, FlowPlan achieves satisfactory results across all 50 trials, demonstrating the versatility of the pipeline. In contrast to the previously mentioned approaches that necessitate re-training or the reconstruction of example sets in novel scenarios with varying constraints, our approach requires only a fundamental understanding of these constraints, which are conveyed through natural language without much effort. Additional experimental videos can be accessed on our project website.

\section{CONCLUSIONS}

This paper presents FlowPlan, a multi-stage workflow that significantly enhances task planning in a zero-shot instruction following scenario. By decomposing the process into interpretable stages\textemdash task information retrieval, language-level reasoning, symbolic-level planning, and logical evaluation\textemdash FlowPlan systematically addresses complex instructions while adhering to operational constraints. It further extracts contextual guidance from instructions to improve instance-level target localization. Experimental results on the ALFRED benchmark demonstrate significant improvements over existing zero-shot methods, with real-world applications validating its versatility in a novel operational context. While FlowPlan bridges the performance gap between zero-shot and data-driven in-context learning approaches, future work could integrate vision-language models for enhanced contextual alignment or leverage open-vocabulary detection for fully off-the-shelf deployment. These directions promise to advance task planning in data-free robotic instruction following systems.

\addtolength{\textheight}{-2.3cm}









\bibliographystyle{ieeetr}



\bibliography{reference}

\begin{thebibliography}{10}

\bibitem{blukis2022persistent}
V.~Blukis, C.~Paxton, D.~Fox, A.~Garg, and Y.~Artzi, ``A persistent spatial semantic representation for high-level natural language instruction execution,'' in {\em Conference on Robot Learning}, pp.~706--717, PMLR, 2022.

\bibitem{min2021film}
S.~Y. Min, D.~S. Chaplot, P.~Ravikumar, Y.~Bisk, and R.~Salakhutdinov, ``Film: Following instructions in language with modular methods,'' {\em arXiv preprint arXiv:2110.07342}, 2021.

\bibitem{inoue2022prompter}
Y.~Inoue and H.~Ohashi, ``Prompter: Utilizing large language model prompting for a data efficient embodied instruction following,'' {\em arXiv preprint arXiv:2211.03267}, 2022.

\bibitem{kim2023context}
B.~Kim, J.~Kim, Y.~Kim, C.~Min, and J.~Choi, ``Context-aware planning and environment-aware memory for instruction following embodied agents,'' in {\em Proceedings of the IEEE/CVF International Conference on Computer Vision}, pp.~10936--10946, 2023.

\bibitem{murray2022following}
M.~Murray and M.~Cakmak, ``Following natural language instructions for household tasks with landmark guided search and reinforced pose adjustment,'' {\em IEEE Robotics and Automation Letters}, vol.~7, no.~3, pp.~6870--6877, 2022.

\bibitem{bhambri2023multi}
S.~Bhambri, B.~Kim, and J.~Choi, ``Multi-level compositional reasoning for interactive instruction following,'' in {\em Proceedings of the AAAI Conference on Artificial Intelligence}, vol.~37, pp.~223--231, 2023.

\bibitem{chen2023robogpt}
Y.~Chen, W.~Cui, Y.~Chen, M.~Tan, X.~Zhang, D.~Zhao, and H.~Wang, ``Robogpt: an intelligent agent of making embodied long-term decisions for daily instruction tasks,'' {\em arXiv preprint arXiv:2311.15649}, 2023.

\bibitem{zhao2024epo}
Q.~Zhao, H.~Fu, C.~Sun, and G.~Konidaris, ``Epo: Hierarchical llm agents with environment preference optimization,'' {\em arXiv preprint arXiv:2408.16090}, 2024.

\bibitem{song2023llm}
C.~H. Song, J.~Wu, C.~Washington, B.~M. Sadler, W.-L. Chao, and Y.~Su, ``Llm-planner: Few-shot grounded planning for embodied agents with large language models,'' in {\em Proceedings of the IEEE/CVF International Conference on Computer Vision}, pp.~2998--3009, 2023.

\bibitem{sarch2024helper}
G.~Sarch, S.~Somani, R.~Kapoor, M.~J. Tarr, and K.~Fragkiadaki, ``Helper-x: A unified instructable embodied agent to tackle four interactive vision-language domains with memory-augmented language models,'' {\em arXiv preprint arXiv:2404.19065}, 2024.

\bibitem{shi2024opex}
H.~Shi, Z.~Sun, X.~Yuan, M.-A. C{\^o}t{\'e}, and B.~Liu, ``Opex: A component-wise analysis of llm-centric agents in embodied instruction following,'' {\em arXiv preprint arXiv:2403.03017}, 2024.

\bibitem{kim2024multi}
T.~Kim, B.~Kim, and J.~Choi, ``Multi-modal grounded planning and efficient replanning for learning embodied agents with a few examples,'' {\em arXiv preprint arXiv:2412.17288}, 2024.

\bibitem{shin2024socratic}
S.~Shin, J.~Kim, G.-C. Kang, B.-T. Zhang, {\em et~al.}, ``Socratic planner: Inquiry-based zero-shot planning for embodied instruction following,'' {\em arXiv preprint arXiv:2404.15190}, 2024.

\bibitem{ridnik2024code}
T.~Ridnik, D.~Kredo, and I.~Friedman, ``Code generation with alphacodium: From prompt engineering to flow engineering,'' {\em arXiv preprint arXiv:2401.08500}, 2024.

\bibitem{aksitov2023rest}
R.~Aksitov, S.~Miryoosefi, Z.~Li, D.~Li, S.~Babayan, K.~Kopparapu, Z.~Fisher, R.~Guo, S.~Prakash, P.~Srinivasan, {\em et~al.}, ``Rest meets react: Self-improvement for multi-step reasoning llm agent,'' {\em arXiv preprint arXiv:2312.10003}, 2023.

\bibitem{li2024survey}
X.~Li, S.~Wang, S.~Zeng, Y.~Wu, and Y.~Yang, ``A survey on llm-based multi-agent systems: workflow, infrastructure, and challenges,'' {\em Vicinagearth}, vol.~1, no.~1, p.~9, 2024.

\bibitem{chen2023llmstate}
S.~Chen, A.~Xiao, and D.~Hsu, ``Llm-state: Expandable state representation for long-horizon task planning in the open world,'' 2023.

\bibitem{rana2023sayplan}
K.~Rana, J.~Haviland, S.~Garg, J.~Abou-Chakra, I.~Reid, and N.~Suenderhauf, ``Sayplan: Grounding large language models using 3d scene graphs for scalable task planning,'' {\em arXiv preprint arXiv:2307.06135}, 2023.

\bibitem{ahn2022can}
M.~Ahn, A.~Brohan, N.~Brown, Y.~Chebotar, O.~Cortes, B.~David, C.~Finn, C.~Fu, K.~Gopalakrishnan, K.~Hausman, {\em et~al.}, ``Do as i can, not as i say: Grounding language in robotic affordances,'' {\em arXiv preprint arXiv:2204.01691}, 2022.

\bibitem{chen2023autotamp}
Y.~Chen, J.~Arkin, Y.~Zhang, N.~Roy, and C.~Fan, ``Autotamp: Autoregressive task and motion planning with llms as translators and checkers,'' {\em arXiv preprint arXiv:2306.06531}, 2023.

\bibitem{brohan2022rt}
A.~Brohan, N.~Brown, J.~Carbajal, Y.~Chebotar, J.~Dabis, C.~Finn, K.~Gopalakrishnan, K.~Hausman, A.~Herzog, J.~Hsu, {\em et~al.}, ``Rt-1: Robotics transformer for real-world control at scale,'' {\em arXiv preprint arXiv:2212.06817}, 2022.

\bibitem{brohan2023rt}
A.~Brohan, N.~Brown, J.~Carbajal, Y.~Chebotar, X.~Chen, K.~Choromanski, T.~Ding, D.~Driess, A.~Dubey, C.~Finn, {\em et~al.}, ``Rt-2: Vision-language-action models transfer web knowledge to robotic control,'' {\em arXiv preprint arXiv:2307.15818}, 2023.

\bibitem{jiang2022vima}
Y.~Jiang, A.~Gupta, Z.~Zhang, G.~Wang, Y.~Dou, Y.~Chen, L.~Fei-Fei, A.~Anandkumar, Y.~Zhu, and L.~Fan, ``Vima: General robot manipulation with multimodal prompts,'' {\em arXiv preprint arXiv:2210.03094}, vol.~2, no.~3, p.~6, 2022.

\bibitem{singh2023progprompt}
I.~Singh, V.~Blukis, A.~Mousavian, A.~Goyal, D.~Xu, J.~Tremblay, D.~Fox, J.~Thomason, and A.~Garg, ``Progprompt: Generating situated robot task plans using large language models,'' in {\em 2023 IEEE International Conference on Robotics and Automation (ICRA)}, pp.~11523--11530, IEEE, 2023.

\bibitem{shridhar2020alfred}
M.~Shridhar, J.~Thomason, D.~Gordon, Y.~Bisk, W.~Han, R.~Mottaghi, L.~Zettlemoyer, and D.~Fox, ``Alfred: A benchmark for interpreting grounded instructions for everyday tasks,'' in {\em Proceedings of the IEEE/CVF conference on computer vision and pattern recognition}, pp.~10740--10749, 2020.

\bibitem{yin2024novel}
Y.-J. Yin, B.-Y. Chen, and B.~Chen, ``A novel llm-based two-stage summarization approach for long dialogues,'' {\em arXiv preprint arXiv:2410.06520}, 2024.

\bibitem{yao2024hdflow}
W.~Yao, H.~Mi, and D.~Yu, ``Hdflow: Enhancing llm complex problem-solving with hybrid thinking and dynamic workflows,'' {\em arXiv preprint arXiv:2409.17433}, 2024.

\bibitem{han2025multi}
Y.~Han and C.~Lyu, ``Multi-stage guided code generation for large language models,'' {\em Engineering Applications of Artificial Intelligence}, vol.~139, p.~109491, 2025.

\bibitem{trirat2024automl}
P.~Trirat, W.~Jeong, and S.~J. Hwang, ``Automl-agent: A multi-agent llm framework for full-pipeline automl,'' {\em arXiv preprint arXiv:2410.02958}, 2024.

\bibitem{grunde2023designing}
M.~Grunde-McLaughlin, M.~S. Lam, R.~Krishna, D.~S. Weld, and J.~Heer, ``Designing llm chains by adapting techniques from crowdsourcing workflows,'' {\em arXiv preprint arXiv:2312.11681}, 2023.

\bibitem{xiao2024flowbench}
R.~Xiao, W.~Ma, K.~Wang, Y.~Wu, J.~Zhao, H.~Wang, F.~Huang, and Y.~Li, ``Flowbench: Revisiting and benchmarking workflow-guided planning for llm-based agents,'' {\em arXiv preprint arXiv:2406.14884}, 2024.

\bibitem{fan2024workflowllm}
S.~Fan, X.~Cong, Y.~Fu, Z.~Zhang, S.~Zhang, Y.~Liu, Y.~Wu, Y.~Lin, Z.~Liu, and M.~Sun, ``Workflowllm: Enhancing workflow orchestration capability of large language models,'' {\em arXiv preprint arXiv:2411.05451}, 2024.

\bibitem{li2024autodcworkflow}
L.~Li, L.~Fang, and V.~I. Torvik, ``Autodcworkflow: Llm-based data cleaning workflow auto-generation and benchmark,'' {\em arXiv preprint arXiv:2412.06724}, 2024.

\bibitem{dorbala2023can}
V.~S. Dorbala, J.~F. Mullen, and D.~Manocha, ``Can an embodied agent find your “cat-shaped mug”? llm-based zero-shot object navigation,'' {\em IEEE Robotics and Automation Letters}, vol.~9, no.~5, pp.~4083--4090, 2023.

\bibitem{chaplot2020object}
D.~S. Chaplot, D.~P. Gandhi, A.~Gupta, and R.~R. Salakhutdinov, ``Object goal navigation using goal-oriented semantic exploration,'' {\em Advances in Neural Information Processing Systems}, vol.~33, pp.~4247--4258, 2020.

\bibitem{ramakrishnan2022poni}
S.~K. Ramakrishnan, D.~S. Chaplot, Z.~Al-Halah, J.~Malik, and K.~Grauman, ``Poni: Potential functions for objectgoal navigation with interaction-free learning,'' in {\em Proceedings of the IEEE/CVF Conference on Computer Vision and Pattern Recognition}, pp.~18890--18900, 2022.

\bibitem{chang2023goat}
M.~Chang, T.~Gervet, M.~Khanna, S.~Yenamandra, D.~Shah, S.~Y. Min, K.~Shah, C.~Paxton, S.~Gupta, D.~Batra, {\em et~al.}, ``Goat: Go to any thing,'' {\em arXiv preprint arXiv:2311.06430}, 2023.

\bibitem{luo2022stubborn}
H.~Luo, A.~Yue, Z.-W. Hong, and P.~Agrawal, ``Stubborn: A strong baseline for indoor object navigation,'' in {\em 2022 IEEE/RSJ International Conference on Intelligent Robots and Systems (IROS)}, pp.~3287--3293, IEEE, 2022.

\bibitem{achiam2023gpt}
J.~Achiam, S.~Adler, S.~Agarwal, L.~Ahmad, I.~Akkaya, F.~L. Aleman, D.~Almeida, J.~Altenschmidt, S.~Altman, S.~Anadkat, {\em et~al.}, ``Gpt-4 technical report,'' {\em arXiv preprint arXiv:2303.08774}, 2023.

\bibitem{kirillov2023segment}
A.~Kirillov, E.~Mintun, N.~Ravi, H.~Mao, C.~Rolland, L.~Gustafson, T.~Xiao, S.~Whitehead, A.~C. Berg, W.-Y. Lo, {\em et~al.}, ``Segment anything,'' in {\em Proceedings of the IEEE/CVF international conference on computer vision}, pp.~4015--4026, 2023.

\bibitem{sundermeyer2021contact}
M.~Sundermeyer, A.~Mousavian, R.~Triebel, and D.~Fox, ``Contact-graspnet: Efficient 6-dof grasp generation in cluttered scenes,'' in {\em 2021 IEEE International Conference on Robotics and Automation (ICRA)}, pp.~13438--13444, IEEE, 2021.

\end{thebibliography}

\end{document}